\pdfoutput=1
\documentclass{article}
\usepackage{spconf,amsmath,graphicx}
\usepackage{cite}
\usepackage{array, multirow}
\usepackage{epsfig} 
\usepackage{amsmath} 
\usepackage{amssymb}  
\usepackage{multirow}
\usepackage{algorithm}
\usepackage{algpseudocode}

\usepackage{amsmath}
\usepackage{amssymb}
\usepackage{amsthm}
\usepackage{bbm,dsfont, mathrsfs}
\usepackage{graphicx,epstopdf}
\usepackage{graphics}
\usepackage{wrapfig}
\usepackage{subcaption}
\usepackage{url}
\usepackage{makecell}
\usepackage{multicol}
\newsavebox{\measurebox}
\makeatother

\makeatletter
\usepackage{amsmath,amssymb,amsthm,amsfonts,dsfont,mathrsfs}
\usepackage{bbm}
\usepackage{cite}
\usepackage{epstopdf}
\usepackage{accents}

\usepackage{multicol}

\newtheorem{theorem}{Theorem}
\newtheorem{lemma}[theorem]{Lemma}

\newcommand{\LRT}[2]{%
  \mathrel{\mathop\gtrless\limits^{#1}_{#2}}%
}
\renewcommand{\leq}{\leqslant} 
\renewcommand{\geq}{\geqslant}

\def\qed{ \hfill $\blacksquare$}  

\DeclareMathOperator{\pr}{\mathds{P}}

\DeclareMathOperator{\argmax}{argmax}

\title{Locally optimal detection of stochastic targeted universal adversarial perturbations}
\name{Amish Goel \hspace{3mm} Pierre Moulin}
\address{Department of Electrical and Computer Engineering,\\ University of Illinois at Urbana-Champaign,\\ Urbana, Illinois 61801}

\begin{document}
%
\maketitle
\begin{abstract}
Deep learning image classifiers are known to be vulnerable to small adversarial perturbations of input images. In this paper, 
we derive the locally optimal generalized likelihood ratio test (LO-GLRT) based detector for detecting stochastic targeted universal adversarial perturbations (UAPs) of the classifier inputs. We also describe a supervised training method to learn the detector's parameters, and demonstrate better performance of the detector compared to other detection methods on several popular image classification datasets. 
\end{abstract}
\begin{keywords}
{Image classification, Neural networks, Locally optimal tests, Universal adversarial perturbations } 
\end{keywords}
\section{Introduction}
\label{sec:intro} 
\begin{figure}[h!]
	\hspace{-0.6cm}
	\includegraphics[clip,width=9.1cm]{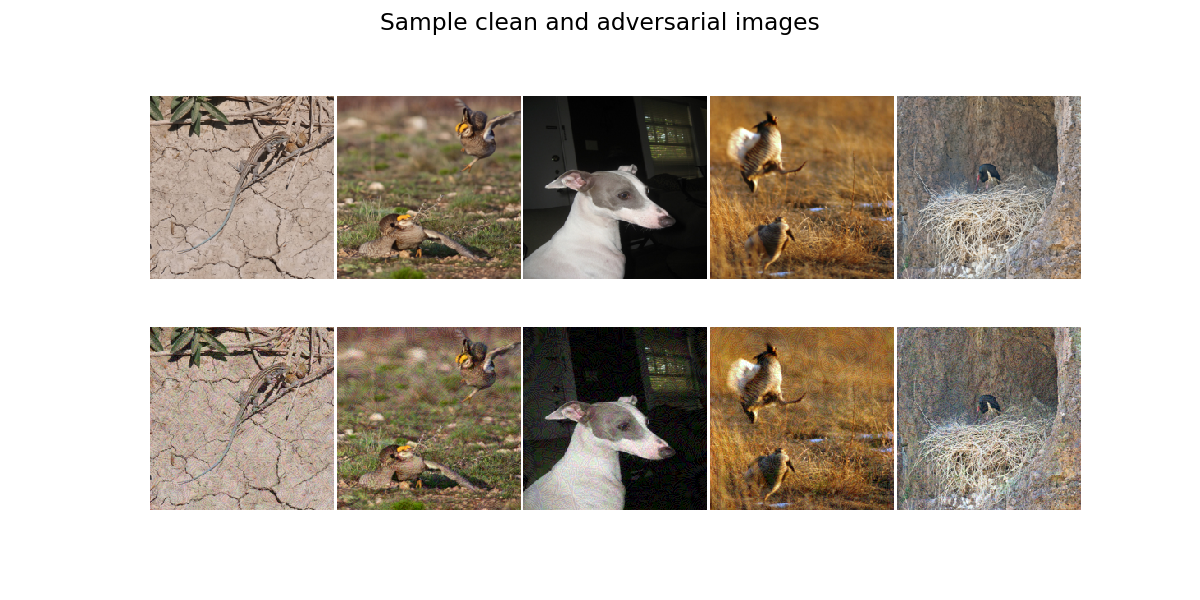}
	\centering
	\caption{(Top): Sample clean images from ImageNet dataset \cite{krizhevsky2012imagenet}. (Bottom): Images corrupted by adding targeted UAPs computed for Resnet-18 classifier \cite{he2016deep}. The UAPs have $\ell_\infty$-norm bound by $\xi = 10$ in the pixel space. }
	\label{figs:images_corrupted}
\end{figure}

Deep neural network (DNN) based classifiers are known to be vulnerable to small carefully crafted   perturbations added to  their inputs. The \emph{adversarial} perturbations can be crafted using a local neighborhood search around an input, such as by moving along the gradient of the classifier's logits \cite{goodfellow2014explaining},  \cite{madry2017towards}. The perturbation can either change the prediction  of the classifier to anything else (in which case it is called non-targeted perturbation), or force the classifier's output to lie in a certain target class (targeted perturbation). Moreover, small universal adversarial perturbations also exist, which can fool a classifier on a significant fraction ($\geq 75$\%) of the inputs    \cite{moosavi2017universal} (see Figure \ref{figs:images_corrupted}). UAPs are a serious concern for many practical applications such as autonomous driving, which rely on DNN based models for their perception. 

Current strategies to defend a classifier against adversarial perturbations fall into two categories: (1) adapting the training algorithms to learn models which are more robust, using for example adversarial training \cite{tramer2017ensemble, cohen2019certified,mummadi2019defending}, (2) detect and possibly rectify the adversarially perturbed inputs \cite{moulin2017locally,goel2018random,metzen2017detecting}. While adversarial training has shown improvement of the DNN against both input dependent and input independent perturbation in images, the improved robustness often come at expense of accuracy on unperturbed inputs \cite{mummadi2019defending}. This robustness-accuracy tradeoff is believed to be fundamental in some of the learning problems \cite{tsipras2018robustness}. The detection methods take a different approach of defence, wherein the defended classifier remains unchanged. The detector extracts discriminative features between the unperturbed and adversarially perturbed inputs, and a simple binary classifier is used on these features to classify the input \cite{metzen2017detecting}, \cite{akhtar2018defense}. 

Most existing detection methods consider input-dependent perturbations. Only a few  discuss input-independent perturbations such as  \cite{akhtar2018defense}, \cite{agarwal2018image}. Akhtar and Liu \emph{et al} \cite{akhtar2018defense} trains a perturbation rectifying network (PRN), which processes the received input image to obtain a rectified image. The difference between the rectified image and the received image is transformed using the Discrete Cosine Transform (DCT), and fed into a binary linear classifier to predict if the input is unperturbed or adversarially perturbed. The detector achieved high classification accuracy against non-targeted UAPs on ImageNet dataset \cite{akhtar2018defense}. However, a drawback of the PRN based detector is its lack of interpretability due to its complex functional form. In contrast, Agarwal \emph{et al.} \cite{agarwal2018image} use a simple interpretable detector that performs principal component analysis (PCA) on an image, and train a support vector machines (SVM) based linear classifier on the PCA score vectors of the image. The detector is shown to achieve low classification error rates for UAPs obtained using Fast Feature Fool \cite{mopuri2017fast} and Moosavi et al's iterative procedure \cite{moosavi2017universal} on face databases. One shortcoming of both the PRN and PCA based detectors is that the detectors are trained and evaluated for only non-targeted UAPs that produces high fooling rates on the classifier. But the output of the classifier against non-targeted UAPs spans only a few labels \cite{moosavi2017universal}. Hence, it is not clear if the detection performance generalizes to more diverse targeted UAPs whose fooling rate might be slightly lower but still significant for many applications. 

In this paper, we consider locally optimal (LO) hypothesis testing for detection of the targeted UAPs in an input to a classifier. LO detectors are more interpretable than other detection methods for small-norm UAPs. LO detectors were earlier described in \cite{moulin2017locally,goel2018random}, for the setting of detecting finitely many  perturbations of the input that are known to the detector. Here, we derive a  locally optimal generalized likelihood ratio test (LO-GLRT) for detecting random targeted UAPs in an input of a classifier. Then we also consider a special case of the test, assuming product multivariate Gaussian distribution on the input. The resulting detector exhibits much simple form than other detectors. We also describe a supervised training method for learning the detector's parameters. We then  evaluate the detector on three key metrics and show that LO-GLRT detector achieves better performance than PRN and PCA based detectors on all the metrics. 


\section{Universal Adversarial Perturbations}
\label{sec:uap}
Let $\mathcal{X} = \mathbb{R}^d$ denote the input space and $\mathcal{Y}$ denote a set of labels. Denote by $(x, y) \in \mathcal{X} \times \mathcal{Y}$ a pair of input and corresponding label drawn according to a joint distribution $\pr_{x,y}$. Also, suppose $\pr_x$ denote the marginal distribution on input $x$. Further, let $f$ denote a classifier and $f(x) \in \mathcal{Y}$ denote the classifier's output for input $x$. Let $H(p,q)$ denotes the cross-entropy of two probability vectors $p$ and $q$, and $\hat{\pi}_{f}(x)$ denotes the conditional output probability vector of the classifier $f$, given an input $x$. Denote by $\mathcal{A}$ an attacker, who chooses a UAP $h \in \mathbb{R}^d$ to fool the  classifier $f$. Also, let $[m]$ denotes the set $\{1,2,\ldots m\}$,
\begin{figure}[t]
\centering
\hspace{-0.5cm}
        \includegraphics[width =9cm]{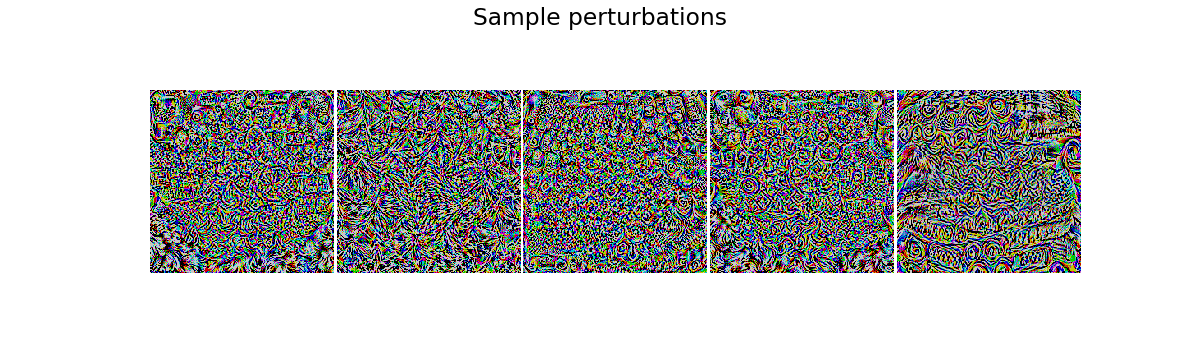}
    \caption{Sample targeted UAPs against GoogleNet classifier. 
    }
    \label{figs:perts_densenet}
\end{figure}

Several recent works discuss methods to find both \emph{targeted}  and \emph{non-targeted} UAPs \cite{moosavi2017universal, hayes2018learning, shafahi2018universal}.  For non-targeted UAPs, Moosavi et al.  \cite{moosavi2017universal} were first to propose an algorithm for generating the perturbations. However, their method is computationally inefficient. Shafahi et al. \cite{shafahi2018universal} use a more tractable approach by solving the optimization problem,
\begin{equation}
\label{eq:shafahi}
\hat{h}_l^* = \argmax_h  \sum_{j=1}^{N} \ell(f(x_j +h), f(x_j)), \quad \|h\|_\infty \leq \xi,
\end{equation}
over $N$ training examples using minibatch stochastic gradient descent (SGD) algorithm, with truncated cross entropy loss function $\ell(\hat{y}, y)$. Other works learn a generative neural network \cite{hayes2018learning},\cite{reddy2018nag} for more efficient sampling of the non-targeted UAPs. The generative network maps a random input to adversarial perturbation, inducing an implicit distribution on the set of  UAPs. To obtain the UAPs for a target class $t$, SGD algorithm is run on a slightly different optimization problem given by, 
\begin{equation}
\label{eq:targeted_empirical}
\min_h \sum_{j=1}^{N}H(\mathbbm{1}_{t},\hat{\pi}_{x_j +h}), \hspace{1mm} \text{subject to } \hspace{1mm} \|h\|_{\infty} \leq \xi,
\end{equation}  
In our experiments, we generate $m$ UAPs per target class by solving the optimization problem in (\ref{eq:targeted_empirical}). We do this efficiently for multiple target classes by learning a $d\times m|\mathcal{Y}|$ dimensional embedding matrix $W$ that maps a pair of target class $t \in \mathcal{Y}$ and an index $k \in [m]$ to a vector in $\mathbb{R}^d$. For each minibatch, we choose the pair $(t,k)$ and substitute $h$ with the column of $W$ corresponding to the pair $(t,k)$. We iterate this process over several minibatches until a desired target classification rate is obtained.  

\begin{figure*}
	\centering
		\centering
		\includegraphics[scale=0.5]{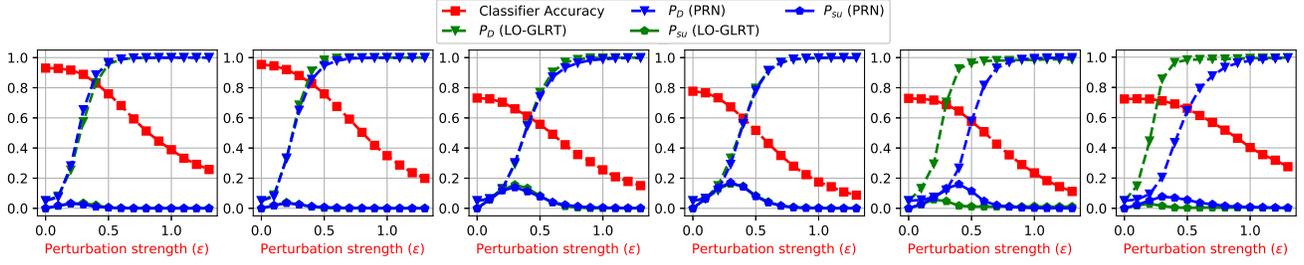}		 
	 \caption{Performance curves of the LO-GLRT detector and PRN detector for detecting UAPs  against (from Left-Right) (a) Resnet-56, (b) Densenet (L=100, k=12) on CIFAR10 dataset, (c) Resnet 164, (d) Densenet (L=100, k=12) on CIFAR100 dataset, and (e) Resnet18, (f) GoogleNet on ILSVRC 2012 dataset.  We use a Gaussian distribution on patches of size $8 \times 8$ in the LO-GLRT detector. Also included classifier's accuracy with increasing perturbation strength for reference.}
 \label{figs:probplot_targeted_all_datasets}
\end{figure*}
\section{Locally optimal detection}
Denote the attacker $\mathcal{A}$'s perturbation distribution by $\nu(h)$. Further, denote the conditional perturbation distribution for a target class $t$ by $\nu(h|t)$. Also, denote the adversarial input by $\tilde{x} = x+\epsilon h$,  where the parameter $\epsilon\geq 0$ models the strength of the perturbation. The perturbed input conditioned on the target class $t$ follows the distribution,
\begin{align*}
\label{eq:targeted_adv}
p_{\epsilon,t}(\tilde{x}) = \int_{\mathbb{R}^d} p(\tilde{x} -\epsilon h) d \nu(h|t), \quad x \in \mathbb{R}^d.
\end{align*}
The detection problem can be formulated as a binary hypothesis testing with composite $H_1$:
\begin{equation}
\label{eq:test_composite}
\left\{\begin{array}{cc}
     \hspace{-1.2cm} H_0 : x \sim p_0,   \\
     H_1 : x \sim p_{\epsilon,t}, \hspace{2mm} t \in \mathcal{Y}. 
\end{array}\right.
\end{equation}
The detector is a decision rule $\delta: \mathbb{R}^d \rightarrow \{0,1\}$,   where $\delta(x) =1$ if the input $x$ is declared adversarial and $\delta(x)=0$ otherwise.  Using the Neyman-Pearson (NP) setup, optimal tests for maximizing $P_D(\epsilon) = P_1\{\delta(x)=1\}$ subject to a false-alarm constraint on the detector, $P_F = P_0\{\delta(x)=1\} \leq \alpha$,  are Likelihood Ratio Tests (LRT) \cite{moulin2019statistical}. For small $\epsilon$, LRT reduces to the  locally optimal (LO) test. For the test in (\ref{eq:test_composite}), the  log-likelihood ratio (LLR) for each target class $t\in \mathcal{Y}$ is given by the following lemma:
\begin{lemma}
Assume $p_0$ is twice continuously differentiable at $x$ and $\nu(h|t)$ has exponential tails $\forall t \in \mathcal{Y}$. Then the LLR, $\hspace{1mm} L_{\epsilon,t}(x) = \frac{p_{\epsilon,t}(x)}{p_0(x)}$, is given as:
\begin{align*}
    \textup{ln}L_{\epsilon,t}(x) &= -\epsilon \bar{h}_t^T\nabla\textup{ln}p_0(x) + O(\epsilon^2) \hspace{2mm} \textup{as}\hspace{1mm} \epsilon \downarrow 0, \nonumber
\end{align*}
\label{lemma_main}
\end{lemma}
\vspace{-0.5cm}
\hspace{-0.5cm}where $\bar{h}_t = \int_{\mathcal{H}(t)} h d\nu(h|t)$ is the mean perturbation for target class $t$.
For multiple target classes, a popular choice is the generalized likelihood ratio test statistic (GLRT) given as,
\begin{equation*}
    L_{G,\epsilon}(x) = \max_{t \in \mathcal{Y}} L_{\epsilon,t}(x).
\end{equation*}
Thus, we can obtain the locally optimal GLRT  statistic as, 
\begin{equation}
\label{eq:test_lod_composite}
T(x) \triangleq \underset{\epsilon \to 0}\lim \frac{1}{\epsilon}\textup{ln}L_{G,\epsilon}(x) = \underset{t \in \mathcal{Y}}\max(-\bar{h}_{t}^T \nabla_x \textup{ln}p_0(x)) \LRT{H_1}{H_0} \tau
\end{equation}
where  $\tau$ is chosen to satisfy $P_F \leq \alpha$. Intuitively, the LO-GLRT detector obtains the gradient of the data's log-density function i.e. $\nabla \textup{ln}p_0(x)$ at the received input, and computes its correlation with the   (negative) mean perturbation vectors for each target class. The set of correlations is max-pooled to obtain the LO-GLRT test statistic.  
\subsection{Detection using constrained distributions}
\label{sec:constraints}
Learning a high dimensional input distribution is generally intractable.  Hence, we consider a constrained probability distribution over the input. In particular, assume the input $x$ can be subdivided into $n$ independent subvectors $x_1, x_2, \cdots x_n \in \mathbb{R}^{d/n}$ that are iid with common  distribution $q_0$. Then $p_0(x) = \prod_i q_0(x_i)$. The perturbation $h$ can likewise be divided into $h_1,\cdots,h_n$ where the subvectors $h_i$ need not be independent. Then, we obtain 
the LO-GLRT statistic as,
\begin{equation}
T(x) = \max_{t \in \mathcal{Y}}\sum_{i=1}^{n}-\left(\bar{h}_{t}\right)_i^T\nabla\textup{ln}q_0(x_i)  \LRT{H_1}{H_0} \tau_n.
\label{eq:lod_patches_composite}
\end{equation}
For images in particular, we simplify the detector using the following constraints: (a) We implement the detector on grayscaled images instead of using all the three channels, (b) we divide the grayscaled-image into $n$ tiles of size $P\times P$ each, where $P$ is some factor of the full image size, and (c) we consider iid multivariate Gaussian distribution over the grayscaled-image tiles. In this case, the gradient over the $i^{\text{th}}$ tile would be given by, $ \nabla \textup{ln}q_0(x_i) = -\Sigma^{-1}(x_i-\mu)$, where $\Sigma, \mu$ denotes the covariance matrix and mean vector of the the Gaussian distribution respectively, and $x_i$ denotes the vectorized $i^{\text{th}}$ tile of the image $x$.
\subsection{Learning the detector parameters}
\label{sec:training_loglrt}
Under the prescribed constraints in section \ref{sec:constraints}, the detector admits the parameters, $\theta = \left\{\Sigma, \mu, \left(\bar{h}_t\right)_{t\in \mathcal{Y}}, \tau\right\}$. 
We learn the parameters in two stages: (a) Unsupervised pretraining: We use maximum likelihood estimation (MLE) to estimate the parameters $\mu, \Sigma$ of the multivariate Gaussian distributions. We use the empirical mean of the targeted perturbations to estimate $\left(\bar{h}_t\right)_{t\in \mathcal{Y}}$, (b) Supervised fine-tuning: We fine tune the parameters estimate from previous stage using supervised binary classification between unperturbed and perturbed images. Specifically, we obtain a set of $N$ images, which contains unperturbed and perturbed images drawn with equal probability, and minimize the binary cross entropy loss given by,
\[
\hat{H}(\theta) =-\sum_{j=1}^{N} y_j \ln \sigma(\beta(x_j;\theta)) + (1-y_j)\ln \left(1-\sigma(\beta(x_j;\theta))\right), \]
where $y_j = 0$ if $x_j$ is unperturbed, $y_j=1$ for perturbed image $x_j$, $\sigma(.)$ denotes the sigmoid function and
\[
\beta\left(x_j;\theta\right) = \max_{t \in \mathcal{Y}}\sum_{i=1}^{n}-\left(\bar{h}_{t}\right)_i^T\Sigma^{-1}\left((x_j)_i-\mu\right)-\tau.
\]
We reparametrize $\Sigma = L L^T$  to ensure that the matrix $\Sigma$ remains positive semi-definite. We minimize the loss $\hat{H}(\theta)$ using a minibatch SGD algorithm, with the parameter values initialized by their estimates from unsupervised pretraining. 
\subsection{Metric for attacker's performance}
We now define a metric to quantify the net success rate of an adversarial perturbation. The success of the perturbations is controlled by two factors:
\begin{enumerate}
\item (Undetectability): $\pr_x[\delta(x+ \epsilon h) = 0]$ is high.
\item (Utility): $\pr_{x}[f(x + \epsilon h) \neq f(x)]$ is high.
\end{enumerate}
Generally, using a small $\epsilon$ makes a perturbation undetectable but a sufficiently high $\epsilon$ is useful for fooling the classifier. For the attacker $\mathcal{A}$ using the perturbation distribution $\nu$,  we can define the probability of successful undetected attack as, 
\begin{align}
\label{prob_su_cond}
P_{\text{su}}(\epsilon) = \pr_{x,h \sim \nu}[f(x+\epsilon h) \neq f(x) \cap  
 \delta(x+\epsilon h) = 0].
\end{align}

\section{Experiments}
\label{sec:expts}
\textbf{Setup}: We evaluate the LO-GLRT detector for detection of UAPs for three popular image classification datasets: CIFAR10, CIFAR100 \cite{krizhevsky2009learning} and ImageNet validation set (ILSVRC 2012) \cite{deng2009imagenet}. The first two datasets contain 60,000 images of size $32 \times 32$, and are typically split into 50,000/10,000 for training/testing respectively. ILSVRC 2012 validation set contains 50,000 variable sized images that are resized into $224 \times 224$ for classification. We use 40,000 of the images for training and the rest 10,000 of the images for evaluation. The three datasets are divided into 10, 100 and 1000 classes respectively for classification. Each image consists of three channels $\{\text{R,G,B}\}$, and each pixel is quantized into 8 bits with pixel values ranging from $0$ to $255$.   \\ 
\textbf{Attack generation}: We generate 50 perturbations for each target class in CIFAR10, 10 perturbations in CIFAR100, and 2 perturbations for ImageNet using the method described in section \ref{sec:uap}. We used fewer perturbations for ImageNet due to high dimension of the images in dataset, which prohibits learning a large embedding matrix $W$. For CIFAR10 and CIFAR100, we use $\xi = 8$, while for ImageNet we used $\xi=10$ to bound the maximum $\ell_\infty$-norm of the perturbations. Some sample UAPs for GoogleNet are shown in Figure \ref{figs:perts_densenet}. \\
\textbf{Detector's training}: Next, we train the LO-GLRT detector, the PRN detector and the PCA based detector against the generated UAPs. For LO-GLRT detector, we use $P=8$ for all three datasets. The training method of the LO-GLRT detector is described in section \ref{sec:training_loglrt}. For the PRN detector, we first train a PRN network to rectify the perturbed images, and then train a SVM classifier to discriminate between the unperturbed and perturbed images. The PRN architecture used in our experiments is same as \cite{akhtar2018defense}. For PCA based detector, we first compute a vector of principal component scores from each grayscaled-image, and then train a SVM classifier on the vector of scores to classify the image into clean and adversarial. For CIFAR10 and CIFAR100, all the 1024 principal components are used, as that gave lowest classification error rate. For ImageNet, we used 5000 principal components that explains $\geq$ 95\% of the data variance. \\
\textbf{Evaluation Metrics}:
We evaluate the detector using the following metrics: (1) classification accuracy on a sample of unperturbed images and adversarially perturbed images drawn with probability half each, (2) detection probability $P_D(\epsilon)$ for varying perturbation strengths with $\alpha=0.05$, and (3) probability of successful undetected attack $P_{\text{su}}(\epsilon)$.  Note that $\epsilon=1$ corresponds to $\xi=8$ for CIFAR10/CIFAR100 and $\xi=10$ for ImageNet. 
\section{Results}
The classification accuracy of all the detectors is shown in the Table \ref{tab:fixed_perturbation}. We observe that PCA  based detector fails to classify the diverse targeted adversarial perturbations accurately. Overall, LO-GLRT detector outperforms the PRN detector in classification accuracy. 
\begin{table}[t]
\centering
\resizebox{\columnwidth}{!}{
\begin{tabular}{|c|c|c|c|c|c|}\hline
\multirow{2}{*}{Dataset} & \multirow{2}{*}{Classifier} & \multicolumn{3}{c}{Detector} \vline\\ \cline{3-5}
&& LO-GLRT & PRN & PCA \\ \cline{1-5}
\multirow{3}{*}{CIFAR10} & \multirow{1}{*}{Resnet56 \cite{he2016deep}} & 0.974 & \bf{0.975} & 0.68  \\ \cline{2-5}
& \multirow{1}{*}{Densenet100 \cite{huang2017densely}} & \bf{0.978} & 0.966 & 0.62  \\ \cline{2-5}
& \multirow{1}{*}{VGG16 \cite{simonyan2014very}} & \bf{0.975} & 0.965 & 0.6  \\ \cline{1-5}
\multirow{3}{*}{CIFAR100} & \multirow{1}{*}{Resnet164  \cite{he2016deep}} & 0.947 & \bf{0.949} & 0.67  \\ \cline{2-5}
& \multirow{1}{*}{Densenet100 \cite{huang2017densely}} & 0.952 & \bf{0.953} & 0.58  \\ \cline{2-5}
& \multirow{1}{*}{WRN-28-10 \cite{zagoruyko2016wide}} & \bf{0.986} & 0.966 & 0.61  \\ \cline{1-5}
\multirow{3}{*}{ImageNet} & \multirow{1}{*}{Resnet18  \cite{he2016deep}} & \bf{0.981} & 0.979 & 0.5  \\ \cline{2-5}
& \multirow{1}{*}{GoogleNet \cite{szegedy2015going}} & \bf{0.968} & 0.963 & 0.5  \\ \cline{2-5}
& \multirow{1}{*}{Alexnet \cite{krizhevsky2012imagenet}} & \bf{0.953} & 0.94 & 0.5  \\ \cline{1-5}
\end{tabular}}
\caption{Classification accuracy of several detectors on a dataset with unperturbed and perturbed images drawn by equal probability. The $\ell_\infty$-norm of the perturbations is bound by $8$ for CIFAR10/CIFAR100, and by $10$ for ImageNet.}
\label{tab:fixed_perturbation}
\end{table}
For $P_D(\epsilon)$ and $P_{\text{su}}(\epsilon)$, we compare LO-GLRT and PRN detectors only, since PCA based detector doesn't perform competitively. The plots for $P_D(\epsilon)$ and $P_{\text{su}}(\epsilon)$ of the two detectors is shown in Figure \ref{figs:probplot_targeted_all_datasets}.  Both the detectors achieve high detection probability for wide range of perturbation strengths, with LO-GLRT detector more robust to smaller perturbation strengths than PRN detector on ImageNet dataset. Further, $P_{\text{su}}(\epsilon)$ of LO-GLRT detector is smaller, or at worst equal to the PRN detector. 

\section{Conclusion}
\label{sec:conclusion}
In this paper, we derived the LO-GLRT for detection of random targeted UAPs added to the input of an image classifier. We also described a supervised training procedure for learning the detector's parameters. We observed high classification accuracy, high detection probability, and low probability of successful undetected attack, and showed better performance of the LO-GLRT detector relative to the PRN detector and superiority against PCA based detectors for three popular image classification datasets.  
\vfill
\pagebreak
\bibliographystyle{IEEEtran}
\bibliography{refs}

\end{document}